\documentclass[pdflatex,sn-mathphys-num]{sn-jnl}

\usepackage{graphicx}%
\usepackage{multirow}%
\usepackage{amsmath,amssymb,amsfonts}%
\usepackage{amsthm}%
\usepackage{mathrsfs}%
\usepackage[title]{appendix}%
\usepackage{xcolor}%
\usepackage{textcomp}%
\usepackage{manyfoot}%
\usepackage{booktabs}%
\usepackage{placeins}
\usepackage{cleveref}
\crefname{figure}{Fig.}{Figs.}\Crefname{figure}{Fig.}{Figs.}
\usepackage{algorithm}%
\usepackage{algorithmicx}%
\usepackage{algpseudocode}%
\usepackage{listings}%
\usepackage{booktabs}
\usepackage{longtable}
\usepackage{pdflscape}
\usepackage{float}
\usepackage{pdfpages}
\usepackage{url}

\theoremstyle{thmstyleone}%
\theoremstyle{thmstyletwo}%

\theoremstyle{thmstylethree}%

\begin{document}

\title[FITTER: Guideline-Grounded Multi-Agent Exercise Prescription from Clinical and Wearable Data]{A Guideline-Grounded Multi-Agent Framework for Personalized Exercise Prescription from Clinical and Wearable Data}


\author[1,3]{\fnm{Xintong} \sur{Li}}\email{xintong.li2@emory.edu}
\author[2,4]{\fnm{Yufen} \sur{Lin}}\email{yufen.lin@emory.edu}
\author[1]{\fnm{Yuhua} \sur{Wu}}\email{gloria.wu@emory.edu}
\author[1]{\fnm{Fengran} \sur{Wang}}\email{fengran.wang@emory.edu}
\author[1]{\fnm{Minxiao} \sur{Wang}}\email{minxiao.wang@emory.edu}
\author[2]{\fnm{Rebecca A.} \sur{Gary}}\email{ragary@emory.edu}
\author[5]{\fnm{Jacob} \sur{Kariuki}}\email{jacob.kariuki@emory.edu}
\author[6]{\fnm{James} \sur{Smith}}\email{j.a.smith6@lancaster.ac.uk}
\author[7]{\fnm{Sandra A.} \sur{Billinger}}\email{sbillinger@kumc.edu}
\author[2]{\fnm{Canhua} \sur{Xiao}}\email{canhua.xiao@emory.edu}
\author[1,3,8]{\fnm{Xiao} \sur{Hu}}\email{xiao.hu@emory.edu}
\author*[1]{\fnm{Runze} \sur{Yan}}\email{runze.yan@emory.edu}

\affil*[1]{\orgdiv{Center for Data Science, Nell Hodgson Woodruff School of Nursing}, \orgname{Emory University}, \orgaddress{\city{Atlanta}, \state{GA}, \country{USA}}}
\affil[2]{\orgdiv{Nell Hodgson Woodruff School of Nursing}, \orgname{Emory University}, \orgaddress{\city{Atlanta}, \state{GA}, \country{USA}}}
\affil[3]{\orgdiv{Department of Computer Science}, \orgname{Emory University}, \orgaddress{\city{Atlanta}, \state{GA}, \country{USA}}}
\affil[4]{\orgdiv{Cancer Prevention and Control Research Program, Winship Cancer Institute}, \orgname{Emory University}, \orgaddress{\city{Atlanta}, \state{GA}, \country{USA}}}
\affil[5]{\orgdiv{Lillian Carter Center for Global Health and Social Responsibility, Nell Hodgson Woodruff School of Nursing}, \orgname{Emory University}, \orgaddress{\city{Atlanta}, \state{GA}, \country{USA}}}
\affil[6]{\orgdiv{Lancaster Medical School}, \orgname{Lancaster University}, \orgaddress{\city{Lancaster}, \country{UK}}}
\affil[7]{\orgdiv{Department of Neurology}, \orgname{University of Kansas Medical Center}, \orgaddress{\city{Kansas City}, \state{KS}, \country{USA}}}
\affil[8]{\orgdiv{Wallace H. Coulter Department of Biomedical Engineering}, \orgname{Georgia Institute of Technology and Emory University}, \orgaddress{\city{Atlanta}, \state{GA}, \country{USA}}}

\abstract{
Translating population-level exercise guidance into individualized prescriptions requires integrating clinical and wearable data—including diagnoses, medications, functional limitations, and baseline activity—with condition-specific recommendations, a process difficult to scale beyond specialist rehabilitation settings. We developed FITTER, a multi-agent framework that integrates these data with condition-specific guidance to generate personalized FITT-VP prescriptions and applies deterministic safety audit and repair to correct rule-checkable violations. We evaluated FITTER using data from 400 All of Us participants (300 cancer, 100 stroke) against five LLM comparators through computational evaluation, ablation, and structured clinician review. FITTER achieved 98.5\% safety-constraint compliance versus 75.8–85.8\% for comparators and showed greater quantitative dosing precision and guideline alignment. Cancer-cohort ablation identified safety audit and repair as the largest contributor to safety-constraint compliance. Clinicians reviewed 80 prescriptions across safety, personalization, instruction concreteness, anticipated effectiveness, and clinical acceptability. FITTER provides a clinician-facing approach integrating clinical and wearable data, guidance, and explicit safety constraints.
} 




\maketitle

\section{Introduction}\label{sec1}

Physical inactivity is a leading modifiable risk factor for chronic disease, premature mortality, and reduced quality of life~\cite{el2024behavioral}. For people living with or recovering from serious illness, particularly cancer and stroke survivors, structured physical activity is associated with improved functional outcomes, reduced recurrence risk, and better psychosocial well-being~\cite{bland2021attention,billinger2014physical,gordon2004physical,rock2020american}. Exercise prescription translates population-level physical activity recommendations into individualized, actionable plans that define what exercise should be performed, at what dose, and how that dose should change over time. The Frequency, Intensity, Time, Type, Volume, and Progression (FITT-VP) framework provides the standard structure for designing and progressively adjusting exercise prescription in clinical practice~\cite{american2013acsm,barisic2011importance,bushman2018developing,campbell2019exercise}. Its broad clinical relevance is reflected in American College of Sports Medicine (ACSM) resources that address exercise interventions across more than 35 health conditions~\cite{thompson2024clinical}. By specifying both the initial exercise dose and its progression, FITT-VP distinguishes a clinical exercise prescription from general advice to become more physically active. 


Despite its established clinical role, translating population-level exercise recommendations into individualized FITT-VP prescriptions remains challenging. An effective prescription must integrate condition-specific guidance with patient-specific information, including diagnoses, medications, functional limitations, baseline activity, personal goals, and practical circumstances~\cite{barisic2011importance,billinger2015does,currier2026american}. These characteristics and clinical risks jointly shape all six FITT-VP dimensions rather than mapping to any single component in isolation. Diagnosis, comorbidities, medications, and functional limitations constrain which exercise types are appropriate, the intensity and duration that can be safely tolerated, the weekly volume that is feasible, and the rate at which the prescription can progress. The FITT-VP dimensions are also interdependent and must be coordinated to maintain clinical appropriateness and internal consistency. Developing such individualized prescriptions therefore requires substantial clinical expertise and time, making the process difficult to scale beyond specialist rehabilitation settings~\cite{billinger2015does,bland2021attention}. This access gap is particularly consequential for cancer and stroke survivors, whose exercise needs vary widely and may be complicated by fall risk, fatigue, impaired mobility, cardiotoxic treatment, neuropathy, and other condition- or treatment-related effects~\cite{billinger2014physical,rock2020american}. A scalable solution must therefore improve not only personalization, but also the timeliness and accessibility with which clinically informed prescriptions can be generated and reviewed.

Digital health provides a potential pathway toward this goal. Mobile applications can improve physical activity engagement when they provide concrete and contextually relevant recommendations~\cite{milne2020mobile,payne2015behavioral}. The growing availability of real-world data linking electronic health records with consumer wearables also creates an opportunity to combine diagnoses, medications, functional characteristics, and longitudinal activity measurements when developing an exercise plan~\cite{khasentino2025personal}. Wearable-derived steps, active minutes, sedentary time, sleep, and physiological measurements can provide an objective representation of baseline behavior that complements clinical and self-reported information. However, existing digital exercise approaches generally do not jointly integrate clinical information, wearable-derived baseline activity, condition-specific guidance, and participant-specific safety constraints~\cite{jones2020checklists,stults2014effects}. Conventional rule-based systems can reliably enforce predefined thresholds but are less suited to interpreting heterogeneous records, retrieving context-dependent evidence, and resolving competing recommendations~\cite{sutton2020overview}. The central challenge is therefore not digital delivery alone, but integrating heterogeneous clinical information with condition-specific guidance while enforcing explicit safety constraints.

Large language models (LLMs) offer a promising foundation for personalized exercise prescription because they can interpret heterogeneous clinical information, synthesize evidence, and generate structured recommendations~\cite{gurses2025role,zhu2024could}. LLM-based systems have shown potential in longitudinal health coaching, rehabilitation planning, and multi-step reasoning over patient data~\cite{khasentino2025personal,li2025care}. Recent work has extended this direction to prompt-based generation of personalized exercise recommendations and FITT-VP prescriptions~\cite{feng2026comparative,li2025exploring}. In one comparative evaluation using 30 clinically informed synthetic participant profiles, four general-purpose LLMs generated all six FITT-VP components, although performance varied substantially across models and prescription dimensions~\cite{feng2026comparative}. These studies demonstrate the feasibility of LLM-generated exercise prescriptions, but they primarily rely on single-call prompting with predefined or synthetic inputs. Real-world application therefore requires a framework that can integrate clinical and wearable data with condition-specific guidance, coordinate interdependent FITT-VP decisions, and enforce participant-specific safety constraints.

To address these gaps, we developed \textbf{FITTER}, a guideline-grounded multi-agent framework for personalized exercise prescription from clinical and wearable data. FITTER models prescription development as a multi-stage clinical decision process rather than a single text-generation task. A structured participant profile integrates diagnoses, medications, functional limitations, and baseline activity, which FITTER combines with condition-specific guidance to generate a personalized FITT-VP prescription. Five sequential LLM-based agents generate Type, Frequency, Intensity and Time, Volume, and Progression, with each decision informed by relevant guidance and the outputs of preceding agents. This structure makes dependencies among FITT-VP components explicit and helps maintain consistency across the prescription. FITTER further combines flexible LLM reasoning with retrieval-augmented generation (RAG), grounding each prescription step in evidence retrieved from the ACSM position stand, the American Heart Association/American Stroke Association (AHA/ASA) scientific statement on exercise for stroke survivors, the American Cancer Society (ACS) guideline, and the American Diabetes Association (ADA) position statement. FITTER also incorporates explicit safety verification. Participant-specific safety constraints are derived from clinical information, medication-related precautions, and an LLM safety review and are provided during prescription generation. After generation, a deterministic safety audit and repair step checks the completed prescription against these constraints, corrects rule-checkable violations, and updates dependent FITT-VP components when necessary. Together, these components enable FITTER to integrate clinical and wearable data with condition-specific guidance and explicit safety constraints to generate clinician-reviewable personalized exercise prescriptions.

We evaluated FITTER using the All of Us Research Program, one of the few population-scale research resources in the United States that links longitudinal electronic health records, participant surveys, physical measurements, medication records, and consumer wearable data at the individual level~\cite{all2019all}. This linked data resource provides clinical and behavioral information needed to evaluate prescription generation from real-world participant profiles. Among 4{,}554 eligible participants with a cancer or stroke diagnosis, we randomly sampled 400 participants, including 300 with cancer and 100 with stroke. FITTER was compared with five LLM-based generation comparators representing direct prompting, step-by-step reasoning, iterative refinement, reasoning-and-action generation, and multi-agent collaboration: Vanilla LLM, Chain-of-Thought (CoT)~\cite{wei2022chain}, Self-Refine~\cite{madaan2023self}, Reasoning and Action (ReAct)~\cite{yao2022react}, and Multi-Agent Collaboration (Collab)~\cite{hong2024metagpt}. Prescriptions were assessed using computational measures of model-assessed safety, exercise-instruction concreteness, and guideline alignment, together with component ablation and structured clinician review. FITTER achieved 98.5\% model-assessed safety-constraint compliance, compared with 75.8–85.8\% for the five comparators, and component ablation showed that deterministic audit and repair made the largest contribution to this improvement. FITTER also showed greater guideline alignment and more precise quantitative specification of exercise dose. Among 80 clinician-reviewed prescriptions, 37 were endorsed without modification, 35 with suggested modification, and 8 were not endorsed. These results indicate that combining guideline retrieval, sequential FITT-VP generation, and explicit safety verification can produce clinician-reviewable exercise prescriptions from real-world clinical and wearable data, while clinician review remains necessary before use.

\section{Results}\label{sec2}

\subsection{Study cohort and prescription generation}

Applying the study cohort definitions to All of Us records identified 4{,}554 participants eligible for evaluation: 3{,}764 with cancer and 790 with stroke. We randomly selected 300 participants from the cancer cohort and 100 from the stroke cohort for model evaluation. The All of Us records served as evaluation inputs; they were not used to train or fine-tune the language models.

FITTER and the five comparator systems---Vanilla LLM, CoT, ReAct, Self-Refine, and Collab---each generated one prescription for every participant in the 400-participant evaluation sample, yielding 2{,}400 prescriptions. Additionally, three modified versions of FITTER were evaluated on the same 300 cancer profiles. These versions removed guideline retrieval, model-assisted safety review, or deterministic audit and repair one at a time and yielded 900 additional prescriptions.

\Cref{tab:cohort} reports the characteristics of the two evaluation samples and their eligible source cohorts. Median age was 66.0 years (interquartile range [IQR], 57.0--72.2) in the cancer sample and 66.5 years (IQR, 56.0--72.0) in the stroke sample. Median body mass index was 28.3 kg/m$^2$ (IQR, 25.0--32.2) and 30.5 kg/m$^2$ (IQR, 25.4--34.1), respectively; women comprised 64.3\% of the cancer sample and 56.0\% of the stroke sample. Median weekly step counts were 44{,}400 for participants with cancer and 35{,}600 for participants with stroke. Median sedentary time was 86.9 and 91.6 hours per week, respectively.

As shown in \cref{tab:cohort}, all absolute standardized mean differences between the cancer evaluation sample and the eligible cancer cohort were below $0.16$. Most differences between the stroke evaluation sample and the eligible stroke cohort were also below $0.20$. The exceptions were resting heart rate ($0.39$), sedentary hours ($0.27$), poor sleep quality ($-0.24$), visual impairment ($-0.22$), and more than one population ($-0.22$). Resting heart rate, which had the largest difference, was available for fewer than half of the eligible stroke cohort. \Cref{tab:cohort} reports simulated goals and practical circumstances separately from observed All of Us variables.

\enlargethispage{3\baselineskip}
Supplementary Figs.~1 and~2 present two simulated cases illustrating the structured inputs and the structure of FITTER prescriptions.

\begin{landscape}
\small
\setlength{\tabcolsep}{2pt}
\renewcommand{\arraystretch}{0.95}

\begin{longtable}{@{}lrrrrrr@{}}
\caption{Characteristics of the eligible cancer and stroke cohorts and the two randomly selected evaluation samples comprising 400 participants. Continuous variables are reported as median [Q1, Q3] and categorical variables as $n$ (\%). Pain, fatigue, general health and sleep variables are self-reported. Standardized mean differences (SMDs) compare each diagnosis-specific evaluation sample with its eligible source cohort; absolute values below $0.10$ indicate very small differences and values below $0.20$ indicate small differences.}
\label{tab:cohort}\\

\toprule
& \multicolumn{3}{c}{Cancer cohort} & \multicolumn{3}{c}{Stroke cohort} \\
\cmidrule(lr){2-4} \cmidrule(lr){5-7}
Characteristic & Full & Evaluation & SMD & Full & Evaluation & SMD \\
& ($n=3{,}764$) & ($n=300$) & & ($n=790$) & ($n=100$) & \\
\midrule
\endfirsthead

\multicolumn{7}{@{}l}{\tablename\ \thetable\ continued from previous page}\\
\toprule
& \multicolumn{3}{c}{Cancer cohort} & \multicolumn{3}{c}{Stroke cohort} \\
\cmidrule(lr){2-4} \cmidrule(lr){5-7}
Characteristic & Full & Evaluation & SMD & Full & Evaluation & SMD \\
& ($n=3{,}764$) & ($n=300$) & & ($n=790$) & ($n=100$) & \\
\midrule
\endhead

\midrule
\multicolumn{7}{r}{Continued on next page}\\
\endfoot

\bottomrule
\endlastfoot

Age, years & 67.0 [58.0, 73.0] & 66.0 [57.0, 72.2] & -0.052 & 67.0 [57.0, 73.0] & 66.5 [56.0, 72.0] & -0.062 \\
Body mass index, kg/m$^{2}$ & 28.1 [24.7, 32.5] & 28.3 [25.0, 32.2] & 0.027 & 28.6 [25.0, 33.8] & 30.5 [25.4, 34.1] & 0.133 \\
Systolic blood pressure, mmHg & 127.0 [119.0, 136.3] & 127.6 [118.5, 138.0] & 0.024 & 126.3 [118.1, 134.0] & 124.5 [120.5, 134.5] & 0.071 \\
Diastolic blood pressure, mmHg & 74.0 [69.0, 79.0] & 73.8 [68.0, 78.9] & -0.044 & 73.1 [68.0, 78.0] & 73.2 [69.4, 77.5] & 0.061 \\
Resting heart rate, bpm & 72.0 [64.7, 79.9] & 72.4 [66.6, 78.8] & -0.018 & 72.1 [66.0, 79.1] & 75.3 [70.8, 82.4] & 0.385 \\
Average pain, past 7 days (0--10) & 2.0 [1.0, 4.0] & 2.0 [1.0, 4.0] & -0.062 & 3.0 [1.0, 5.0] & 3.0 [1.0, 5.0] & 0.018 \\

\addlinespace[2pt]
\multicolumn{7}{@{}l}{\textit{Average fatigue level, past 7 days}} \\
\quad Mild & 1,675 (44.5) & 135 (45.0) & 0.010 & 307 (38.9) & 45 (45.0) & 0.124 \\
\quad Moderate & 980 (26.0) & 66 (22.0) & -0.094 & 267 (33.8) & 31 (31.0) & -0.060 \\
\quad Severe & 246 (6.5) & $<$20 & -0.022 & 69 (8.7) & $<$20 & -0.027 \\
\quad Very Severe & 35 (0.9) & $<$20 & -0.030 & $<$20 & $<$20 & -0.092 \\
\quad Unknown & 828 (22.0) & 79 (26.3) & 0.101 & 130 (16.5) & $<$20 & -0.040 \\
Weekly step count, $\times 10^{3}$ & 43.5 [28.6, 63.4] & 44.4 [28.7, 65.6] & 0.029 & 35.9 [23.4, 56.0] & 35.6 [20.4, 55.2] & -0.098 \\
Very active minutes per week & 60.0 [9.0, 176.0] & 67.5 [9.2, 185.5] & 0.063 & 29.5 [0.0, 118.0] & 28.0 [2.0, 106.8] & -0.073 \\
Fairly active minutes per week & 68.0 [23.0, 143.0] & 66.0 [23.5, 153.5] & 0.033 & 47.0 [8.8, 116.2] & 49.0 [14.0, 111.2] & -0.057 \\
Lightly active hours per week & 22.8 [17.7, 28.5] & 23.4 [17.9, 29.2] & 0.081 & 21.7 [16.2, 27.1] & 21.6 [15.6, 25.8] & -0.023 \\
Sedentary hours per week & 86.8 [74.9, 98.8] & 86.9 [76.0, 98.1] & 0.084 & 89.7 [75.6, 102.2] & 91.6 [82.3, 102.1] & 0.271 \\
Minutes asleep per night & 395.0 [335.1, 436.3] & 393.6 [336.9, 433.2] & 0.061 & 384.9 [308.8, 434.7] & 382.5 [326.4, 425.1] & 0.166 \\
Hemoglobin, g/L & 133.0 [123.0, 142.5] & 134.8 [123.4, 146.0] & 0.060 & 132.7 [121.2, 142.8] & 133.0 [113.8, 143.1] & -0.157 \\
Platelet count, $\times 10^{9}$/L & 232.0 [190.2, 278.5] & 237.5 [203.2, 288.6] & 0.129 & 230.5 [189.8, 281.1] & 224.2 [192.9, 267.2] & -0.172 \\
Glucose, mg/dL & 99.8 [91.0, 114.0] & 99.0 [89.0, 110.5] & -0.120 & 101.0 [90.5, 115.7] & 103.0 [92.8, 115.3] & 0.012 \\

\addlinespace[2pt]
\multicolumn{7}{@{}l}{\textit{Sex}} \\
\quad Female & 2,228 (59.2) & 193 (64.3) & 0.106 & 490 (62.0) & 56 (56.0) & -0.123 \\
\quad Male & 1,492 (39.6) & 103 (34.3) & -0.110 & 289 (36.6) & 42 (42.0) & 0.111 \\
\quad Other, prefer not to answer, or skipped & 44 (1.2) & $<$20 & 0.015 & $<$20 & $<$20 & 0.047 \\

\addlinespace[2pt]
\multicolumn{7}{@{}l}{\textit{Race}} \\
\quad White & 3,123 (83.0) & 252 (84.0) & 0.028 & 611 (77.3) & 80 (80.0) & 0.065 \\
\quad Black or African American & 180 (4.8) & $<$20 & -0.073 & 71 (9.0) & $<$20 & -0.035 \\
\quad Asian & 55 (1.5) & $<$20 & -0.011 & $<$20 & $<$20 & -0.047 \\
\quad More than one population & 142 (3.8) & $<$20 & 0.044 & 36 (4.6) & $<$20 & -0.218 \\
\quad Unknown & 264 (7.0) & 20 (6.7) & -0.014 & 60 (7.6) & $<$20 & 0.085 \\

\addlinespace[2pt]
\multicolumn{7}{@{}l}{\textit{Ethnicity}} \\
\quad Not Hispanic or Latino & 3,461 (92.0) & 281 (93.7) & 0.066 & 717 (90.8) & 91 (91.0) & 0.008 \\
\quad Hispanic or Latino & 209 (5.6) & $<$20 & -0.040 & 50 (6.3) & $<$20 & -0.014 \\
\quad Unknown & 94 (2.5) & $<$20 & -0.058 & 23 (2.9) & $<$20 & 0.005 \\

\addlinespace[2pt]
\multicolumn{7}{@{}l}{\textit{ADL limitation}} \\
\quad Completely & 2,537 (67.4) & 207 (69.0) & 0.034 & 425 (53.8) & 56 (56.0) & 0.044 \\
\quad Mostly & 594 (15.8) & 44 (14.7) & -0.031 & 159 (20.1) & $<$20 & -0.054 \\
\quad Moderately & 388 (10.3) & 32 (10.7) & 0.012 & 131 (16.6) & $<$20 & 0.011 \\
\quad A Little & 210 (5.6) & $<$20 & -0.011 & 66 (8.4) & $<$20 & 0.023 \\
\quad Not At All & $<$20 & $<$20 & -0.080 & $<$20 & $<$20 & -0.143 \\
\quad Unknown & 23 (0.6) & $<$20 & -0.041 & $<$20 & $<$20 & -0.050 \\
Documented fall risk & 96 (2.6) & $<$20 & 0.046 & $<$20 & $<$20 & -0.176 \\
Recent chemotherapy & 231 (6.1) & $<$20 & -0.006 & $<$20 & $<$20 & -0.011 \\

\addlinespace[2pt]
\multicolumn{7}{@{}l}{\textit{Difficulty walking or climbing stairs}} \\
\quad Yes & 562 (14.9) & 47 (15.7) & 0.020 & 189 (23.9) & 22 (22.0) & -0.046 \\
\quad No & 3,156 (83.8) & 249 (83.0) & -0.023 & 590 (74.7) & 76 (76.0) & 0.031 \\
\quad Unknown & 46 (1.2) & $<$20 & 0.010 & $<$20 & $<$20 & 0.047 \\

\addlinespace[2pt]
\multicolumn{7}{@{}l}{\textit{Difficulty concentrating}} \\
\quad Yes & 398 (10.6) & 30 (10.0) & -0.019 & 142 (18.0) & 21 (21.0) & 0.076 \\
\quad No & 3,313 (88.0) & 266 (88.7) & 0.020 & 634 (80.3) & 79 (79.0) & -0.031 \\
\quad Unknown & 53 (1.4) & $<$20 & -0.006 & $<$20 & $<$20 & -0.190 \\

\addlinespace[2pt]
\multicolumn{7}{@{}l}{\textit{Blindness or serious visual impairment}} \\
\quad Yes & 103 (2.7) & $<$20 & 0.087 & 37 (4.7) & $<$20 & -0.223 \\
\quad No & 3,617 (96.1) & 281 (93.7) & -0.110 & 741 (93.8) & 96 (96.0) & 0.100 \\
\quad Unknown & 44 (1.2) & $<$20 & 0.067 & $<$20 & $<$20 & 0.100 \\

\addlinespace[2pt]
\multicolumn{7}{@{}l}{\textit{General physical health}} \\
\quad Excellent & 337 (9.0) & 42 (14.0) & 0.159 & 48 (6.1) & $<$20 & -0.095 \\
\quad Very Good & 1,311 (34.8) & 108 (36.0) & 0.024 & 185 (23.4) & 30 (30.0) & 0.149 \\
\quad Good & 1,299 (34.5) & 103 (34.3) & -0.004 & 304 (38.5) & 39 (39.0) & 0.011 \\
\quad Fair & 680 (18.1) & 39 (13.0) & -0.140 & 205 (25.9) & 23 (23.0) & -0.069 \\
\quad Poor & 122 (3.2) & $<$20 & -0.034 & 45 (5.7) & $<$20 & -0.132 \\
\quad Unknown & $<$20 & $<$20 & -0.089 & $<$20 & $<$20 & 0.075 \\

\addlinespace[2pt]
\multicolumn{7}{@{}l}{\textit{General mental health}} \\
\quad Excellent & 995 (26.4) & 88 (29.3) & 0.065 & 171 (21.6) & 24 (24.0) & 0.056 \\
\quad Very Good & 1,465 (38.9) & 121 (40.3) & 0.029 & 285 (36.1) & 34 (34.0) & -0.044 \\
\quad Good & 890 (23.6) & 68 (22.7) & -0.023 & 202 (25.6) & 31 (31.0) & 0.121 \\
\quad Fair & 327 (8.7) & 21 (7.0) & -0.063 & 101 (12.8) & $<$20 & -0.122 \\
\quad Poor & 69 (1.8) & $<$20 & -0.105 & 28 (3.5) & $<$20 & -0.094 \\
\quad Unknown & $<$20 & $<$20 & -0.098 & $<$20 & $<$20 & -0.087 \\

\addlinespace[2pt]
\multicolumn{7}{@{}l}{\textit{Sleep quality}} \\
\quad Excellent & 408 (10.8) & 41 (13.7) & 0.086 & 92 (11.6) & $<$20 & 0.153 \\
\quad Good & 1,531 (40.7) & 128 (42.7) & 0.040 & 337 (42.7) & 50 (50.0) & 0.148 \\
\quad Fair & 346 (9.2) & 20 (6.7) & -0.094 & 70 (8.9) & $<$20 & -0.069 \\
\quad Poor & 1,479 (39.3) & 111 (37.0) & -0.047 & 291 (36.8) & 26 (26.0) & -0.235 \\

\addlinespace[3pt]
\midrule
\multicolumn{7}{@{}l}{\textit{Simulated scenario context (not All of Us measurements)}} \\

\addlinespace[2pt]
\multicolumn{7}{@{}l}{\textit{Psychological readiness}} \\
\quad High & 1,253 (33.3) & 89 (29.7) & -0.078 & 209 (26.5) & 31 (31.0) & 0.101 \\
\quad Moderate & 1,938 (51.5) & 154 (51.3) & -0.003 & 439 (55.6) & 56 (56.0) & 0.009 \\
\quad Low & 573 (15.2) & 57 (19.0) & 0.100 & 142 (18.0) & $<$20 & -0.138 \\

\addlinespace[2pt]
\multicolumn{7}{@{}l}{\textit{Stress level}} \\
\quad High & 1,509 (40.1) & 134 (44.7) & 0.093 & 352 (44.6) & 37 (37.0) & -0.154 \\
\quad Moderate & 1,534 (40.8) & 115 (38.3) & -0.050 & 341 (43.2) & 53 (53.0) & 0.198 \\
\quad Low & 721 (19.2) & 51 (17.0) & -0.056 & 97 (12.3) & $<$20 & -0.072 \\

\addlinespace[2pt]
\multicolumn{7}{@{}l}{\textit{Social context}} \\
\quad Solo & 1,749 (46.5) & 140 (46.7) & 0.004 & 344 (43.5) & 50 (50.0) & 0.130 \\
\quad Caregiver/Family & 688 (18.3) & 63 (21.0) & 0.069 & 210 (26.6) & 22 (22.0) & -0.107 \\
\quad Friends & 645 (17.1) & 37 (12.3) & -0.136 & 90 (11.4) & $<$20 & 0.019 \\
\quad Clinician-Supervised & 405 (10.8) & 39 (13.0) & 0.069 & 99 (12.5) & $<$20 & -0.150 \\
\quad Group Class/Club & 125 (3.3) & $<$20 & 0.019 & 22 (2.8) & $<$20 & 0.013 \\
\quad Team/Coach & 152 (4.0) & $<$20 & -0.037 & 25 (3.2) & $<$20 & 0.093 \\

\end{longtable}
\end{landscape}

\subsection{Evaluation results}

The evaluation addressed six dimensions of prescription quality: safety, instruction concreteness, personalization, anticipated effectiveness, guideline alignment, and clinical acceptability. Computational evaluation assessed model-assessed safety-constraint compliance, instruction concreteness, and guideline alignment across FITTER and the five comparator systems. Furthermore, clinician review assessed safety, instruction concreteness, personalization, anticipated effectiveness, and clinical acceptability for 80 FITTER prescriptions. The clinician-review sample included 32 cancer prescriptions and 48 stroke prescriptions.

\subsubsection{Safety}
As shown in \cref{fig:objective}a, FITTER satisfied all specified safety constraints in 394 of 400 prescriptions (98.5\%): 297 of 300 cancer prescriptions (99.0\%) and 97 of 100 stroke prescriptions (97.0\%). The corresponding percentages were 75.8\% for Vanilla LLM, 84.2\% for CoT, 82.8\% for ReAct, 85.8\% for Self-Refine, and 77.8\% for Collab. Across the 400 matched participant profiles, FITTER met all specified safety constraints while the comparator did not in 55 to 95 cases, depending on the comparator, whereas the reverse occurred in four to six cases; full discordant-pair counts, risk differences, and Holm-adjusted McNemar tests are given in \cref{tab:systems}. 

Clinicians rated the same construct directly. Asked whether the recommended activities were free from safety concerns, including harmful instructions, contraindicated movements, unsafe intensity or duration, and potential interactions with the participant's diagnoses, medications, or functional status (Supplementary Note~1), reviewers answered yes for 51 of the 80 prescriptions, partial for 27, and no for 2, giving a normalized safety score of 80.6\% (95\% confidence interval, 74.4--86.2; \cref{fig:subjective}). The clinician figure is 18 percentage points below the model-assessed rate. The two measures are not interchangeable: the model assessor scores each plan against a checklist derived from the structured record, whereas clinicians could draw on information the record did not represent (\cref{methods}). We therefore interpret 98.5\% as conformance to a record-derived safety specification rather than as an estimate of clinical safety, and report the two evaluations side by side in Supplementary Note~5.

\subsubsection{Instruction concreteness}

\begin{figure}[!htbp]
  \centering
  \includegraphics[width=\linewidth]{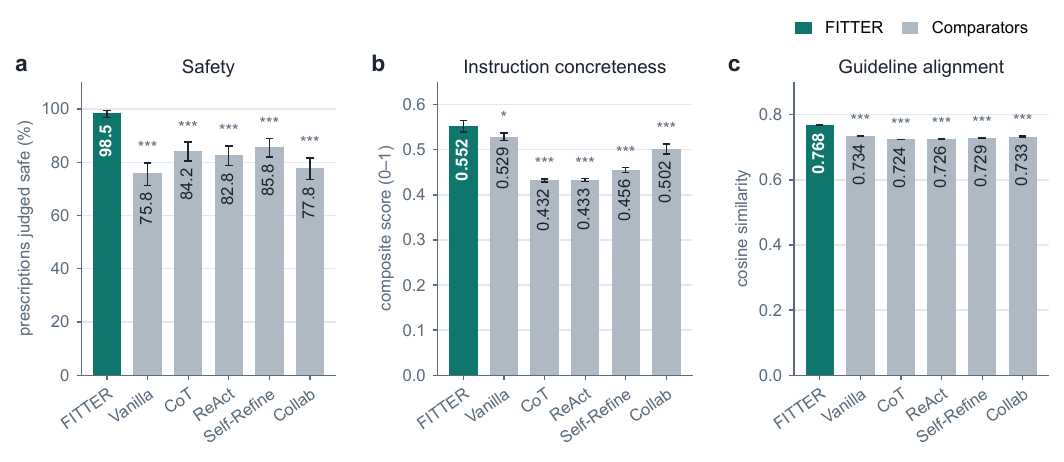}
  \caption{Computational evaluation across the 400 samples. \textbf{a}, Proportion of prescriptions with no detected safety-constraint violation by an independent model assessor. \textbf{b}, Instruction-concreteness composite (0--1). \textbf{c}, Guideline alignment, the embedding (cosine) similarity between each FITT-VP component and the guideline passages, averaged across components to give one value per prescription. FITTER is compared with five LLM comparators. Error bars show 95\% confidence intervals. Asterisks indicate Holm-adjusted significance versus FITTER ($*P < 0.05; ***P < 0.001)$}
  \label{fig:objective}
\end{figure}

\begin{figure}[!htbp]
  \centering
  \includegraphics[width=0.6\linewidth]{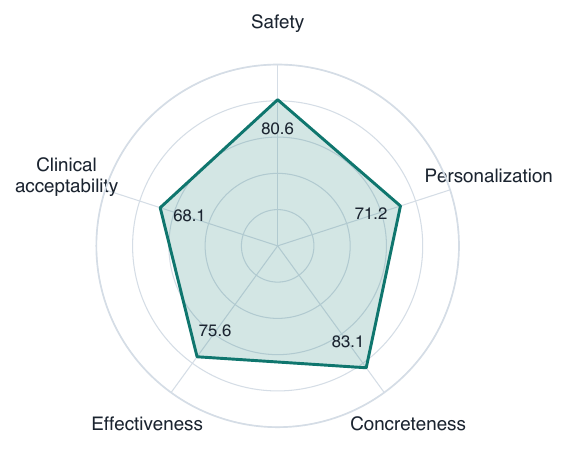}
  \caption{Clinician evaluation of 80 FITTER prescriptions. Each prescription was rated by one clinician reviewer on five aspects using a yes/partial/no scale. Values are mean normalized scores on a 0--100 scale, with yes, partial, and no assigned values of 100, 50, and 0, respectively.}
  \label{fig:subjective}
\end{figure}

Instruction concreteness is a 0-1 composite measuring the extent to which a prescription contains quantitative FITT-VP specifications, exact session timing, distinct named activities, and concrete language; higher values indicate more specific, more directly actionable instructions. As shown in \cref{fig:objective}b, the mean instruction-concreteness score was $0.552$ for FITTER, compared with $0.529$ for Vanilla LLM, $0.432$ for CoT, $0.433$ for ReAct, $0.456$ for Self-Refine, and $0.502$ for Collab. FITTER exceeded CoT, ReAct, Self-Refine, and Collab by $0.120$, $0.119$, $0.096$, and $0.050$, respectively (Holm-adjusted $p < 10^{-7}$; matched-pairs rank-biserial correlations ranging from $0.32$ to $0.85$). The difference from Vanilla LLM was $0.023$ (95\% confidence interval, $0.009$--$0.038$; Holm-adjusted $p = 0.046$; rank-biserial correlation $0.11$). As presented in \cref{tab:systems}, the complete pairwise results support these comparisons.

FITTER scored highest on quantitative specifications ($0.641$) and exact session timing ($0.492$), the two components capturing exercise dose; the other components assess sentence-level lexical concreteness and distinct named activities. The component scores and weighting sensitivity qualify the primary computational result: it concerns dosing precision, not every dimension of instruction concreteness. Full component-level results and weighting sensitivity analyses are reported in Supplementary Table~2. Clinicians were asked whether the instructions were specific, actionable, and feasible to implement, with clearly defined parameters such as named exercises, sets and repetitions, duration, and intensity targets (Supplementary Note~1). They answered yes for 59 of the 80 prescriptions, partial for 15, and no for 6 (83.1\%; 95\% CI, 76.2--89.4), with scores of 90.6\% for cancer and 78.1\% for stroke (Supplementary Table~3).

\subsubsection{Guideline alignment}

As shown in \cref{fig:objective}c, FITTER had a mean guideline-alignment score of $0.768$. The comparator scores were $0.734$ for Vanilla LLM, $0.724$ for CoT, $0.726$ for ReAct, $0.729$ for Self-Refine, and $0.733$ for Collab. FITTER exceeded these systems by $0.033$, $0.044$, $0.042$, $0.039$, and $0.035$, respectively. The overall Friedman test gave $\chi^2 = 1055.1$ ($p = 7.2\times10^{-226}$); each FITTER--comparator difference remained significant after Holm correction ($p < 10^{-64}$), with matched-pairs rank-biserial correlations of at least $0.98$.

Relative performance among the comparator strategies varied across endpoints. Among the comparators, Self-Refine showed the highest model-assessed safety (85.8\%), followed by CoT (84.2\%), ReAct (82.8\%), Collab (77.8\%), and Vanilla (75.8\%). In contrast, Vanilla ranked highest among comparators for instruction concreteness (0.529) and guideline alignment (0.734), followed by Collab on both measures (0.502 and 0.733, respectively). Thus, no comparator consistently dominated across prescription-quality dimensions, and more elaborate reasoning, refinement, or collaboration strategies did not correspond to uniformly better performance.

\begin{table}[!htbp]
\caption{Cross-system comparison across 400 matched participant profiles. Each metric was first tested with a Friedman or Cochran omnibus test across the six systems, then FITTER was compared with each comparator, and the five resulting $P$ values were Holm-adjusted. Differences are FITTER minus comparator, so positive values favor FITTER. The pairwise measure is the matched-pairs rank-biserial correlation for the continuous metrics and the discordant-pair counts (FITTER safe/comparator safe) for the binary safety outcome. Confidence intervals are from 10{,}000 bootstrap resamples of the paired differences.}
\label{tab:systems}
\centering
\scriptsize
\setlength{\tabcolsep}{4pt}
\begin{tabular}{@{}lrrrr@{}}
\toprule
System & Score & Difference [95\% CI] & Holm-adjusted $P$ & Pairwise measure \\
\midrule
\addlinespace[2pt]
\multicolumn{5}{@{}l}{\textit{Safety, proportion judged safe}; Cochran $Q=128.9$, $p=4.1\times10^{-26}$} \\
FITTER & 98.5\% & reference & --- & --- \\
\quad vs Vanilla & 75.8\% & +22.8 pp & $6.2\times10^{-23}$ & 95/4 \\
\quad vs CoT & 84.2\% & +14.2 pp & $7.8\times10^{-14}$ & 61/4 \\
\quad vs ReAct & 82.8\% & +15.7 pp & $3.5\times10^{-14}$ & 69/6 \\
\quad vs Self-Refine & 85.8\% & +12.7 pp & $1.7\times10^{-12}$ & 55/4 \\
\quad vs Collab & 77.8\% & +20.8 pp & $9.0\times10^{-21}$ & 87/4 \\
\addlinespace[4pt]
\multicolumn{5}{@{}l}{\textit{Instruction concreteness (0--1)}; Friedman $\chi^2=706.7$, $p=1.7\times10^{-150}$} \\
FITTER & 0.552 & reference & --- & --- \\
\quad vs Vanilla & 0.529 & +0.023 [0.009, 0.038] & 0.046 & 0.11 \\
\quad vs CoT & 0.432 & +0.120 [0.107, 0.133] & $2.4\times10^{-48}$ & 0.85 \\
\quad vs ReAct & 0.433 & +0.119 [0.106, 0.132] & $1.1\times10^{-47}$ & 0.84 \\
\quad vs Self-Refine & 0.456 & +0.096 [0.083, 0.110] & $3.9\times10^{-31}$ & 0.67 \\
\quad vs Collab & 0.502 & +0.050 [0.033, 0.067] & $3.9\times10^{-8}$ & 0.32 \\
\addlinespace[4pt]
\multicolumn{5}{@{}l}{\textit{Guideline alignment (cosine similarity)}; Friedman $\chi^2=1055.1$, $p=7.2\times10^{-226}$} \\
FITTER & 0.768 & reference & --- & --- \\
\quad vs Vanilla & 0.734 & +0.033 [0.032, 0.035] & $1.6\times10^{-66}$ & 1.00 \\
\quad vs CoT & 0.724 & +0.044 [0.042, 0.045] & $1.6\times10^{-66}$ & 1.00 \\
\quad vs ReAct & 0.726 & +0.042 [0.040, 0.043] & $1.6\times10^{-66}$ & 1.00 \\
\quad vs Self-Refine & 0.729 & +0.039 [0.038, 0.040] & $1.6\times10^{-66}$ & 1.00 \\
\quad vs Collab & 0.733 & +0.035 [0.033, 0.036] & $5.7\times10^{-65}$ & 0.98 \\
\addlinespace[2pt]
\bottomrule
\end{tabular}
\end{table}
\FloatBarrier

\subsubsection{Clinical acceptability}

Across the 80 clinician-reviewed FITTER prescriptions, the clinical-acceptability score was 68.1\% (95\% confidence interval, 60.6--75.6). Reviewers accepted 37 prescriptions as written, accepted 35 with a suggested change, and did not accept 8. The score was 73.4\% among the 32 cancer prescriptions and 64.6\% among the 48 stroke prescriptions.

\FloatBarrier

\subsubsection{Clinician comments}

As presented in \cref{tab:themes}, 55 reviews included at least one free-text comment. The two most frequent themes, each present in 19 reviews, were a missing guideline-recommended exercise component and limited use of equipment access or the stated exercise goal. Other comments concerned how well the starting dose matched measured activity ($n=14$), limited detail for non-aerobic exercise ($n=13$), missing functional information ($n=13$), and clinical considerations outside the structured input ($n=12$).

\begin{table}[!htbp]
\caption{Themes raised in at least ten of the 55 reviews with free-text comments. A review could contribute to more than one theme. Representative comments are reproduced verbatim apart from minor spelling corrections.}
\label{tab:themes}
\centering
\footnotesize
\setlength{\tabcolsep}{5pt}
\begin{tabular}{@{}p{4.6cm}rp{6.8cm}@{}}
\toprule
Theme & Reviews & Representative comment \\
\midrule
Additional guideline-recommended exercise components suggested & 19 & ``Only includes cardiorespiratory activity. Other types of exercise (e.g., strength) should be included as per guidelines.'' \\
\addlinespace[3pt]
Equipment access or stated goal not fully reflected in the plan & 19 & ``The patient does not have access to a recumbent cycle.'' \\
\addlinespace[3pt]
Starting dose could be better aligned with measured baseline activity & 14 & ``The patient is already fairly active, the starting intensity and dose is too low.'' \\
\addlinespace[3pt]
Less detail provided for non-aerobic than aerobic components & 13 & ``No specific exercises, sets, or reps for strength exercise. RPE may not be appropriate for strength exercises.'' \\
\addlinespace[3pt]
Additional functional detail requested to assess the plan & 13 & ``I would need more information on the patient's level of disability and functional ability.'' \\
\addlinespace[3pt]
Clinical considerations outside the structured input & 12 & ``Would be concerned about fentanyl use around exercise time, could be a fall risk.'' \\
\bottomrule
\end{tabular}
\end{table}
\FloatBarrier

\subsubsection{Ablation analysis}

\Cref{tab:ablation} reports the ablation on the same 300 cancer profiles, for full FITTER and for each of the three variants. Removing any single component reduced the proportion of prescriptions with no detected safety-constraint violation from $99.0\%$ for the full system to $82.0\%$ without guideline retrieval, to $76.3\%$ without the model-assisted safety review, and to $62.0\%$ without the deterministic audit and repair loop. That loop was therefore the single largest contributor to safety, a 37-percentage-point (pp) drop when removed, followed by the model-assisted safety review (22.7 points) and retrieval (17 points). Each removal was statistically significant (Cochran's Q across the four conditions, $p < 10^{-29}$; pairwise McNemar versus the full system, Holm-adjusted $p < 10^{-13}$). All four conditions were scored by the same judge under a single prompt, so the ordering among them is internally consistent (\cref{tab:ablation}). Notably, the largest contribution comes from the deterministic audit and repair step rather than from either language-model component, indicating that the safety gain rests primarily on rule-based correction of the generated plan.

\begin{table}[!htbp]
\caption{Ablation analysis using 300 cancer participant profiles. Full FITTER and three ablation variants were assessed by the percentage of prescriptions meeting all specified safety constraints. Each variant removed one component, and all four conditions used the same participant profiles and model assessor. Differences are ablation variant minus full FITTER\@, in percentage points (pp). Cochran $Q=137.9$ on 3 degrees of freedom, $p=1.1\times10^{-29}$; exact McNemar tests against full FITTER are Holm-adjusted.}
\label{tab:ablation}
\centering
\footnotesize
\begin{tabular}{@{}lrrr@{}}
\toprule
Condition & Judged safe & Difference & Holm $p$ \\
\midrule
Full FITTER & 99.0\% & reference & --- \\
\quad No guideline retrieval & 82.0\% & -17.0 pp & $8.6\times10^{-14}$ \\
\quad No model-assisted safety review & 76.3\% & -22.7 pp & $2.2\times10^{-18}$ \\
\quad No deterministic audit and repair & 62.0\% & -37.0 pp & $9.6\times10^{-31}$ \\
\bottomrule
\end{tabular}
\end{table}
\FloatBarrier

\section{Discussion}\label{sec3}

This study addresses the challenge of translating population-level exercise recommendations into individualized FITT-VP prescriptions for people with complex clinical characteristics. FITTER integrates patient-specific clinical information, wearable-derived activity, and condition-specific guidance through guideline retrieval, sequential FITT-VP generation, and explicit safety verification. The evaluation yielded three principal findings. First, FITTER achieved higher safety-constraint compliance than the five comparator systems, with ablation identifying deterministic safety audit and repair as the largest contributor to this improvement. Second, FITTER produced more precise quantitative exercise dosing and stronger guideline alignment, while the relative ranking among comparator strategies varied by endpoint: Vanilla was the strongest comparator for instruction concreteness and guideline alignment but the weakest for safety. Third, structured clinician review provided complementary assessment of safety, personalization, instruction concreteness, anticipated effectiveness, and clinical acceptability while identifying specific opportunities for further improvement.

The ablation study showed that guideline retrieval, LLM safety review, and deterministic safety audit and repair made complementary contributions to model-assessed safety-constraint compliance. Removing these components reduced the proportion of prescriptions with no detected safety-constraint violation by 17.0, 22.7, and 37.0 percentage points, respectively. The largest decrease occurred when deterministic audit and repair were removed, indicating that this step provided additional protection beyond generation-time reasoning. Self-Refine asks the model to critique and revise its own output, whereas FITTER checks the completed prescription against an explicit participant-specific constraint set. Prior work has identified limitations in relying on language models to recognize and correct their own errors~\cite{huang2024large,kamoi2024can}. The present findings therefore support a division of roles in which the model generates candidate recommendations and explicit constraints govern requirements that can be stated and checked directly~\cite{omar2025multi,kresevic2024optimization}. This approach is well suited to linked FITT-VP quantities such as frequency, intensity, time, and volume because a correction to one field can trigger consistent updates to dependent fields. Decisions that cannot be expressed as explicit rules still require clinical judgment.

The relative performance of the comparator systems further supports the rationale for FITTER’s domain-specific workflow. Vanilla performed best among the comparators for instruction concreteness and guideline alignment but showed the lowest safety performance, whereas Self-Refine achieved the highest safety among the comparators without matching Vanilla on the other computational endpoints. Thus, increasing generic reasoning complexity through step-by-step reasoning, self-refinement, or multi-agent collaboration did not consistently improve exercise-prescription quality. In contrast, FITTER structures LLM reasoning around the clinical dependencies among FITT-VP decisions and combines this task-specific workflow with condition-specific guidance and explicit safety verification. These findings suggest that reliable exercise-prescription generation depends less on adding general-purpose reasoning complexity than on organizing LLM reasoning around the structure and constraints of the clinical task. 

The computational endpoints require distinct interpretations. The instruction-concreteness score combines quantitative FITT-VP specification, exact session timing, distinct named activities, and lexical concreteness. FITTER performed best on quantitative specification and exact session timing, the components most directly related to exercise dose, while alternative weightings changed the system ordering. The result therefore supports greater dosing precision rather than a general claim that every aspect of the instructions was more concrete. Clinician comments identified a separate property not captured by this measure: some reviewers requested clearer directions for performing activities such as tandem stance or resistance-band exercises. Procedural adequacy and dosing precision should therefore be measured separately. 

Structured clinician review provided complementary evidence on personalization, anticipated effectiveness, and clinical acceptability. Scores for clinical acceptability were 73.4\% for the 32 cancer prescriptions and 64.6\% for the 48 stroke prescriptions; scores for anticipated effectiveness were 95.3\% and 62.5\%, respectively. Reviewers assigned all 12 no ratings for anticipated effectiveness to stroke prescriptions. Their comments suggest two direct extensions to the framework. First, a component-completeness check could verify that the final prescription includes every exercise component recommended for the diagnosis. Second, a profile-consistency check could test whether selected activities reflect equipment access, exercise goals, and wearable-derived activity. These checks would address incomplete use of information already available in the participant profile. Other comments requested more detailed functional information or identified considerations recorded outside the standardized profile; such missing information cannot be recovered by output checks and requires clinician input. The observed omissions may also reflect the capacity of the Qwen3.5-9B backbone rather than the framework itself. Because the study evaluated one backbone, framework effects and model-capacity effects cannot be disentangled.

The appropriate clinical role for FITTER follows from the boundary between rule-checkable requirements and decisions that depend on information or judgment outside the structured profile. To maximize safety, FITTER should serve as clinician-facing decision support, with a clinician reviewing and approving each prescription before use. Prospective evaluation of such a system would follow reporting standards developed for the early-stage clinical assessment of AI-based decision support~\cite{vasey2022reporting}. In the present review, clinicians accepted 37 prescriptions as written and 35 with a suggested modification, suggesting a workflow in which FITTER prepares a structured draft and the clinician focuses on adaptation and approval. Such a workflow may be useful where clinicians have limited time to assemble individualized prescriptions from patient characteristics, medications, wearable-derived activity, and guideline recommendations~\cite{ramsey2022exercise,schmitz2019exercise,moncion2020barriers,fini2021address}. The implementation with an open-weight backbone inside the All of Us Researcher Workbench also shows that guideline retrieval, a medication knowledge base, and explicit safety constraints can operate without exporting participant-level inputs. Related work in personalized nutrition has combined language-model generation with explicit constraints for drug--nutrient interactions~\cite{wu2026nutriorion}, suggesting that this design may extend to other prescriptive tasks with structured doses and identifiable constraints.

The present findings are bounded by the model and deployment environment. Available memory and throughput in the All of Us Researcher Workbench limited the size of models that could run locally, while restrictions on exporting participant-level data precluded evaluation through commercial LLM APIs. The study therefore used one open-weight Qwen3.5-9B backbone. A cross-backbone evaluation using larger open-weight models and, where data governance permits, commercial models could help determine whether the clinician-identified omissions reflect model capacity or framework design. Future evaluation should assess robustness across model backbones and decoding settings and quantify prescription stability across repeated generations. The evaluation framework can be extended in three further directions. First, the guideline corpus can be expanded beyond the four included sources, and embedding-based alignment can be paired with direct checks of prescribed parameters against guideline ranges. Second, the ablation study can be repeated for both diagnosis groups and expanded beyond model-assessed safety to test medication screening, sequential FITT-VP generation, and other framework components across all endpoints. Third, future structured clinician review can compare systems directly through blinded assessment, assign multiple reviewers to overlapping prescriptions, and use a prespecified coding framework for free-text comments. This design would permit cross-system clinical comparisons, inter-rater agreement estimates, and more reproducible thematic analysis.

Clinical generalizability requires prospective evaluation with more representative inputs and care settings. The present cohort required linked wearable data, so it may not represent people without consumer activity trackers~\cite{carrier2026demographic,patten2026all,nagappan2024patterns}. Future studies should test alternative estimates of baseline activity when wearable data are unavailable and evaluate FITTER in external cancer and stroke cohorts. Baseline activity was also classified using uniform step-count thresholds, whereas step counts decline with age and differ by sex, so a demographically adjusted classification may set starting doses more appropriately for older or less mobile participants. Exercise goals, psychological readiness, stress, social setting, and equipment access should be collected directly from participants rather than assigned as scenario variables. An input-validation step should also identify implausible wearable values before generation. Finally, prospective workflow studies should deliver the prescriptions under clinician supervision and measure review time, acceptability, adherence, tolerability, changes in physical activity and function, and clinical outcomes. These studies would assess both practical utility and realized treatment effects.

In summary, FITTER translated linked clinical and wearable data together with condition-specific guidance into individualized FITT-VP prescriptions for 400 participants with cancer and stroke, satisfying record-derived safety constraints more often than five LLM-based comparators and specifying exercise dose more precisely. Deterministic audit and repair contributed most to this safety advantage, indicating that explicit, rule-checkable constraints complement language-model reasoning. Clinicians accepted most reviewed prescriptions as written or with modification, and their comments identified omitted exercise components, incomplete use of equipment and goals, and clinical information outside the structured profile as the main targets for improvement. FITTER is therefore best positioned as clinician-facing decision support that prepares a structured draft for clinician review, pending prospective evaluation of its safety, usability, and effect on physical activity and clinical outcomes.

\section{Methods}\label{methods}

\begin{figure}[!htbp]
  \centering
  \includegraphics[width=1\linewidth]{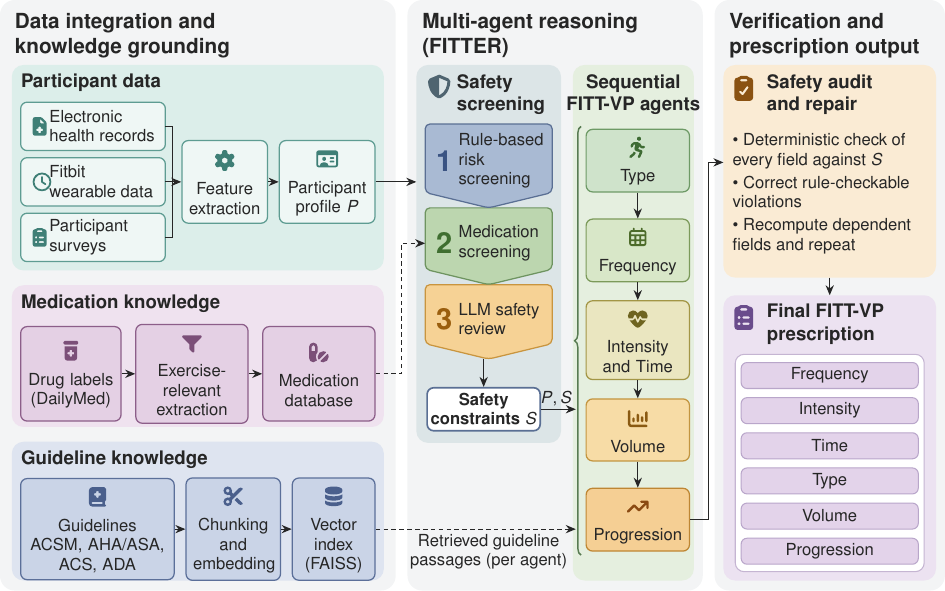}
  \caption{FITTER workflow. Clinical, functional, medication, and wearable data form a structured participant profile. Pre-generation screening produces safety constraints; five sequential agents then generate Type, Frequency, Intensity and Time, Volume, and Progression using retrieved guideline passages. After generation, the complete FITT-VP prescription is checked against those constraints; detected violations are corrected, producing the final structured prescription.}
  \label{fig:pipline}
\end{figure}

As shown in \cref{fig:pipline}, FITTER generates and audits each prescription through a staged workflow. Using existing participant records, we evaluated FITTER on 400 participants from the All of Us Research Program, including 300 with cancer and 100 with stroke. Comparisons with five LLM-based systems tested whether FITTER's structured workflow improved prescription quality relative to alternative generation designs, while component ablations examined which parts of the framework contributed to the observed differences. Evaluation combined computational measures with structured clinician review across six dimensions of prescription quality.

\FloatBarrier

\subsection{Participant cohort}

The All of Us Research Program links electronic health records, participant surveys, physical measurements, medication records, and Fitbit activity data~\cite{all2019all}. We identified 4{,}554 eligible participants with cancer ($n=3{,}764$) or stroke ($n=790$), defined from OMOP standard condition concepts and their descendants; Supplementary Table~1 lists the concept sets. We randomly sampled 300 participants with cancer and 100 with stroke, and every generation system received the same 400 records. Each participant was represented by a standardized profile integrating observed clinical and wearable data from the All of Us Research Program. Observed inputs included demographics; primary diagnosis and selected comorbidities; current medications; vital signs and available laboratory measurements; wearable-derived baseline activity; self-reported symptoms, functional status, general health, sleep, and lifestyle factors. Wearable measures included weekly step count, sedentary time, and minutes at different activity levels, while functional measures included mobility limitations, fall risk, pain, and fatigue. Self-reported measures were derived from All of Us participant surveys rather than clinician assessments. To represent contextual factors relevant to exercise prescription but not available as observed All of Us variables, we augmented each profile with simulated exercise goals, psychological readiness, stress, social setting, and equipment access. These simulated scenario variables were clearly distinguished from and kept separate from observed All of Us measures in cohort descriptions and analyses. All analyses were conducted inside the All of Us Researcher Workbench in accordance with the program’s applicable data-use policies and agreements.

\subsection{Wearable-derived activity level}\label{sec:app_activity}

FITTER classified baseline activity from weekly wearable totals. Participants were categorized as sedentary when they accumulated fewer than 35{,}000 steps or more than 4{,}500 sedentary minutes, and as moderately active when they accumulated at least 70{,}000 steps and 150 minutes of moderate-to-vigorous activity. The step-count bands correspond to fewer than 5{,}000 and at least 10{,}000 steps per day, thresholds established for classifying adult activity~\cite{tudor2004many}, and the 150-minute threshold follows the guideline minimum for moderate-to-vigorous activity~\cite{garber2011american}. All others were categorized as low active. The criteria were applied in this order so that the sedentary classification took precedence. Baseline activity informed the starting dose and rate of progression.

\subsection{FITTER framework}

FITTER comprises three components: evidence sources and retrieval; sequential FITT-VP generation; and safety screening with deterministic audit and repair. During sequential generation, each agent receives the decisions produced by all preceding agents.

\subsubsection{Evidence sources and retrieval}\label{subsec:rag}

Exercise prescriptions for people with cancer or stroke must reconcile general training principles with diagnosis- and comorbidity-specific recommendations. FITTER therefore drew on four sources: the American College of Sports Medicine position stand~\cite{garber2011american}, the AHA/ASA scientific statement on physical activity and exercise for stroke survivors~\cite{billinger2014physical}, the American Cancer Society guideline~\cite{rock2020american}, and the American Diabetes Association position statement~\cite{colberg2016physical}. These four documents differ in formal designation: one is a clinical practice guideline, one a scientific statement, and two are position stands or statements. We refer to them collectively as the guidance corpus, and name the retrieval and alignment components after it for brevity rather than to imply that every source is a clinical practice guideline. Within this framework, FITTER retrieved evidence separately for each FITT-VP decision using the participant's diagnosis, functional or safety concerns, and exercise goal, rather than retrieving the same evidence for the full prescription. Limits on repeated passages prevented one source or section from dominating the evidence supplied to an agent.

The four guidance documents named above were divided into overlapping 350-word passages with a 60-word overlap, yielding 289 passages: 57 from ACSM, 56 from ADA, 90 from ACS, and 86 from AHA/ASA. Passage length, overlap, and the per-source retrieval caps were fixed a priori from pilot inspection of the corpus and were not tuned against any evaluation outcome. Passages were encoded with \texttt{bge-base-en-\allowbreak v1.5} and indexed for similarity retrieval with Facebook AI Similarity Search (FAISS)~\cite{xiao2024c,johnson2019billion}.

For each agent, each of the three decision-specific queries described above retrieved its three nearest passages; baseline activity was also available during query planning. After duplicate passages were merged, FITTER retained at most eight passages, with no more than three from one guideline and two from one section. When available, at least one passage represented each query.

Medication effects can alter safe exercise intensity, modality, and monitoring even when exercise guidelines do not address a participant's drug regimen. FITTER therefore supplemented guideline retrieval with a medication knowledge base derived from FDA Structured Product Labels through DailyMed~\cite{national2014dailymed}. The knowledge base linked each active ingredient to predefined exercise precautions, allowing FITTER to apply the same medication rules to every prescription.

Warnings, adverse reactions, and drug interactions were assigned to seven exercise-related categories: cardiotoxicity, peripheral neuropathy, fall or sedation risk, QT prolongation, bleeding, immunosuppression, and bone-density loss. RxNorm medication names were normalized to base ingredients and matched against a fixed version of the knowledge base throughout the evaluation.

\subsubsection{Sequential FITT-VP generation}

The FITT-VP decisions are interdependent: activity type constrains feasible frequency, intensity, and duration; frequency and duration determine weekly volume; and progression depends on the completed starting prescription. FITTER therefore generated the decisions in the order Type, Frequency, Intensity and Time, Volume, and Progression, so that each agent could condition its decision on all upstream choices rather than generate potentially conflicting elements independently. An expert team that included physical therapists and specialists in stroke and cancer care co-designed this sequence and the prompts used by all five agents. Let $P$ denote the participant profile, $S$ the pre-generation safety constraints, $R_k$ the guideline evidence retrieved for agent $k$, $\tau_k$ its output schema, $\mathcal{A}_k$ the agent itself, and $a_k$ the structured output it returns. \Cref{tab:agents} lists the decision assigned to each agent and the output passed downstream; the five agents operate sequentially as follows:

\begin{equation}
a_k = \mathcal{A}_k\big(P,\ S,\ R_k,\ a_{<k},\ \tau_k\big),
\qquad 
a_{<k} = (a_1,\dots,a_{k-1}),
\qquad 
k = 1,\dots,5,
\label{eq:pipeline}
\end{equation}

\begin{table}[htbp]
\caption{Sequential FITTER agents and their structured outputs.}
\label{tab:agents}
\centering
\small
\setlength{\tabcolsep}{5pt}
\begin{tabular}{@{}p{0.23\linewidth}p{0.75\linewidth}@{}}
\toprule
Agent & Decision and downstream output \\
\midrule
Type & Selects aerobic, resistance, flexibility, and neuromotor components and names the prescribed activities. \\
Frequency & Sets initial and target sessions per week for each included component. \\
Intensity and Time & Sets initial and target RPE ranges, talk-test descriptions, session durations, and stop criteria. \\
Volume & Calculates initial and target weekly minutes for each component and for the full prescription. \\
Progression & Converts the prescription into six two-week blocks with progression, safety, and regression rules. \\
\bottomrule
\end{tabular}
\end{table}

The same expert team defined the eligible activity set by reviewing the 2024 Adult Compendium of Physical Activities~\cite{herrmann20242024} and removing activities it judged unsuitable for people with cancer or stroke. The Type agent selected from this set and included aerobic exercise unless contraindicated. The Frequency agent specified initial and target sessions per week, after which the Intensity and Time agent assigned minutes per session, Borg RPE 0--10 targets, corresponding talk-test descriptions, and heart-rate targets when validated physiological data supported their use. The Volume agent multiplied sessions per week by minutes per session for each component and summed across components to obtain initial and target weekly minutes. The Progression agent then organized the prescription into six two-week blocks, increasing duration before frequency and frequency before intensity; if symptoms worsened or recovery was poor, regression rules reduced total load by 20--30\% for one week.

\subsubsection{Safety screening, audit, and repair}\label{subsec:safety}

FITTER constructed a participant-specific set of safety constraints before generation and checked the completed prescription against those constraints afterward. Potential concerns came from three sources: the participant profile, including health conditions and functional limitations; medication intake, screened through the medication knowledge base; and an LLM safety review. The LLM reviewed the complete standardized participant profile together with the concerns detected by the profile and medication screens to identify potential risks that the predefined screens had missed. The LLM safety review could add concerns but could not remove a concern detected by the first two screens or weaken the constraint mapped from it. A standardized mapping converted every concern into one or more actionable constraints, whose union formed the constraint set $S$. Each constraint identified the source concern, the affected FITT-VP fields, and the required limit, prohibition, or action. Thus, $S$ could specify an intensity ceiling, a prohibited activity, a frequency or duration limit, a volume or progression restriction, or a stop criterion. Every agent received $S$ together with the participant profile.

After generation, FITTER checked every FITT-VP component against all applicable constraints in $S$ and verified consistency among dependent fields. When a violation was detected, FITTER corrected the affected field, recomputed dependent FITT-VP components, and reapplied all relevant constraints. This process was repeated until no rule-checkable violations remained or no further deterministic correction was possible.

\textbf{Profile-derived concerns.} The fixed profile screen evaluated health conditions, vital signs, symptoms, fall risk, and activities-of-daily-living limitations. An explicit fall-risk flag or a severe activities-of-daily-living limitation activated the fall-risk concern. The detected concerns were also combined into a cumulative clinical risk score, which recorded overall pre-generation risk alongside the individual constraints. Neurological, fall, and cardiovascular risk each contributed 2 points, and metabolic risk, respiratory risk, and elevated vital signs each contributed 1 point; total scores of 0--1, 2--4, and $\geq 5$ were classified as low, moderate, and high risk, respectively.

\textbf{Concern-to-constraint mapping.} The standardized mapping prohibited high-fall-risk activities when fall risk was present, capped intensity at RPE~6 under a non-vigorous restriction, required confirmation of cardiac function before vigorous exercise under cardiotoxicity, prohibited vigorous exercise under QT prolongation, and excluded balance-demanding activities under medication-related peripheral neuropathy. LLM-identified concerns used the same mapping. \Cref{tab:safety-constraints} summarizes the constraint classes and the fields governed by each class.

\begin{table}[htbp]
\caption{Standardized safety-constraint records stored in $S$ and the FITT-VP fields governed by each class.}
\label{tab:safety-constraints}
\centering
\small
\setlength{\tabcolsep}{5pt}
\begin{tabular}{@{}p{0.25\linewidth}p{0.45\linewidth}p{0.30\linewidth}@{}}
\toprule
Constraint class & Actionable content & FITT-VP fields checked \\
\midrule
Intensity limit & Permitted RPE or heart-rate range, including a prohibition on vigorous exercise & Intensity; Progression \\
Activity restriction & Prohibited exercise component or named activity & Type; Progression \\
Frequency, time, or volume limit & Bounds on sessions per week, minutes per session, or weekly minutes & Frequency; Time; Volume; Progression \\
Progression restriction & Permitted increases, required holds, or regression conditions & Progression; Volume \\
Stop criterion & Symptom or vital-sign condition requiring exercise to stop & Intensity and Time; Progression \\
\bottomrule
\end{tabular}
\end{table}

\textbf{Deterministic audit and repair.} After all five agents completed the prescription, FITTER evaluated every Frequency, Intensity, Time, Type, Volume, and Progression field against all applicable safety constraints and checked dependencies among these fields. For each detected violation, FITTER identified the affected field and applied the correction specified by the corresponding constraint. Changes to Type or Intensity triggered updates to affected progression blocks, whereas changes to Frequency or Time triggered recalculation of weekly Volume. FITTER then reapplied all relevant safety and progression constraints to the revised prescription. This audit–repair cycle continued until no rule-checkable violations remained or no further deterministic correction was possible. FITTER recorded the original value, revised value, triggering constraint, and repair for every modified field.

\subsection{Comparator systems and ablations}

All five comparator systems used the same language-model backbone, decoding settings, participant profile, and target schema as FITTER.

\begin{enumerate}
  \item \textbf{Vanilla LLM (Vanilla).} A single-pass direct prompt generated the complete FITT-VP prescription using the same participant profile and target schema as the other systems.
  \item \textbf{Chain-of-Thought (CoT).} The model reasoned step by step through the FITT-VP components before producing the structured prescription~\cite{wei2022chain}.
  \item \textbf{Self-Refine.} The model generated an initial prescription, critiqued it, and revised it~\cite{madaan2023self}.
  \item \textbf{Reasoning and Action (ReAct).} The prompt interleaved reasoning with component-level generation~\cite{yao2022react}.
  \item \textbf{Multi-Agent Collaboration (Collab).} Concurrent component agents exchanged feedback over two peer-review rounds before their outputs were combined~\cite{hong2024metagpt}.
\end{enumerate}

The ablation study used the same 300 cancer participants in all four conditions. The three ablations removed: (i) guideline retrieval; (ii) the model-assisted safety review, while retaining deterministic and medication screening; or (iii) deterministic audit and repair, while retaining all pre-generation checks. Thus, every ablation differed from full FITTER by one component and used identical participant inputs.

\subsection{Evaluation framework}

The evaluation addressed six dimensions of generated FITT-VP prescription quality: safety, instruction concreteness, personalization, anticipated effectiveness, guideline alignment, and clinical acceptability. No single measurement method could represent all six constructs. We therefore combined quantitative computational assessment with structured clinician review and selected the measurement approach according to the construct. The computational measures applied the same prespecified procedures to outputs from all six systems, providing standardized quantitative comparisons; clinician review supplied expert judgment for properties that required interpretation of the participant profile and the prescription as a whole. Safety and instruction concreteness received both types of assessment; personalization, anticipated effectiveness, and clinical acceptability were clinician-rated; and guideline alignment was computed from prescription and guideline embeddings. For the structured clinician review, we randomly selected 80 FITTER prescriptions, including 32 for participants with cancer and 48 for participants with stroke, for evaluation by four reviewers. The panel comprised one physical therapist with stroke rehabilitation expertise, two oncology nurse scientists, and one exercise scientist from academic clinical and rehabilitation settings in the United States and the United Kingdom. Supplementary Note~1 provides the clinician evaluation survey. Reviewers were asked to provide a free-text comment whenever they selected partial or no and could comment after any item. Supplementary Note~4 reports pooled and diagnosis-specific ratings in Supplementary Table~3.

\paragraph{Safety}

Safety was defined as compatibility between the prescription and the participant's clinical, functional, and medication constraints. For the computational assessment, the participant-specific constraint set $S$ constructed in \cref{subsec:safety} served as the safety checklist. Because $S$ contained the applicable limits, prohibitions, restrictions, and stop criteria derived from the participant profile, medication intake, and model-assisted safety review, it provided an individualized standard against which prescriptions from all six systems could be evaluated. A GPT-5-family model distinct from the Qwen3.5-9B generation backbone evaluated each generated prescription against this derived safety checklist~\cite{openai2026gpt54mini}. The evaluator received the generated prescription and safety checklist rather than the underlying All of Us clinical and wearable record. We defined \emph{model-assessed safety} as the proportion of prescriptions with no detected checklist violation. Clinicians also rated the safety of FITTER prescriptions through structured expert review. Supplementary Note~5 explains how the two safety endpoints were interpreted.

\paragraph{Instruction concreteness}

Instruction concreteness assessed whether a prescription gave specific, actionable directions about what activity to perform, how often and for how long to perform it, and how the dose should progress. A deterministic text analysis measured how frequently quantitative FITT-VP specifications and exact session-timing statements appeared, counted distinct named activities, and combined these features with sentence-level lexical concreteness derived from the Brysbaert norms~\cite{brysbaert2014concreteness}. After scaling each component to $[0,1]$, we calculated the prescription-level concreteness score. Clinicians separately judged whether the complete instructions were specific and actionable. Supplementary Note~3 reports its components and sensitivity analyses under equal and numeric-only weights (Supplementary Table~2).

\paragraph{Personalization}

Personalization assessed whether the activities, dose, and progression formed a coherent response to the participant's diagnosis, wearable-derived activity, medications, and exercise goals. This construct concerns the fit between a profile and a prescription, so clinicians reviewed both together rather than scoring isolated features of the prescription text. Reviewers judged whether the FITT-VP decisions responded appropriately to the participant's circumstances, allowing the evaluation to distinguish substantive adaptation from simply mentioning profile details. The endpoint therefore represents expert-perceived fit to the constructed study profile.

\paragraph{Effectiveness}

Effectiveness was evaluated prospectively as \emph{anticipated effectiveness}: whether the prescribed exercise dose and progression were likely to achieve the intended goals. Clinician assessment was well suited to this construct because it required the FITT-VP components to be considered together in the participant's clinical context and in relation to the stated goals. Reviewers assessed consistency with evidence-based guidance and the clinical appropriateness of the FITT-VP parameters. This endpoint captures the expected benefit of the proposed prescription; the retrospective design did not measure adherence, functional change, or realized treatment effects.

\paragraph{Guideline alignment}

Guideline alignment assessed how closely a generated prescription matched the language and meaning of the clinical guidance used by FITTER. Because the same recommendation can be expressed in different words, exact word matching would miss valid paraphrases. We therefore used embedding similarity to compare each FITT-VP component with the guideline passages and averaged the component scores to obtain one value per prescription. A higher score indicated closer textual and semantic similarity to the guideline corpus; it did not by itself establish that the prescription was clinically appropriate for a specific participant. \Cref{subsec:rag} describes the guideline corpus and embedding model.

\paragraph{Clinical acceptability}

Clinical acceptability assessed whether a clinical expert would accept the prescription for use with that participant under real-world clinical conditions. Reviewers compared the generated prescription with the prescription they would consider appropriate for the same participant and rated their willingness to use the generated prescription in clinical practice.

\subsubsection{Analysis of clinician comments}

Reviewers provided free-text comments alongside their ratings; a comment was required whenever a reviewer selected \textit{partial} or \textit{no} on an item and was optional otherwise. Comments made under all five rating aspects were pooled for analysis. After all ratings had been collected, one investigator reviewed all comments and grouped them into recurring themes without a prespecified codebook and without reference to the quantitative ratings. Themes were not mutually exclusive: a single review could contribute to more than one, and reported counts are numbers of reviews contributing to a theme rather than numbers of individual comments. Because thematic grouping was performed by a single investigator without independent double-coding, inter-rater reliability was not estimated. The analysis was intended as a descriptive summary of the issues raised by reviewers rather than a formal qualitative analysis.

\subsection{Statistical analyses}\label{sec:stats}

Continuous cohort variables are reported as medians with first and third quartiles, and categorical variables as counts and percentages. Standardized mean differences were used to assess sample--cohort balance because evaluation samples were randomly drawn from their source cohorts and standardized differences are not driven by sample size.

Each system is summarized descriptively over the full 400-prescription sample by the proportion of prescriptions with no detected safety-constraint violation, the mean instruction-concreteness composite and its four sub-scores, and the mean guideline-alignment similarity; the proportion judged safe is additionally reported for the cancer ($n=300$) and stroke ($n=100$) strata separately.

The five clinician-rated endpoints were summarized descriptively for the 80 reviewed FITTER prescriptions; Supplementary Table~3 reports the pooled and diagnosis-specific ratings. Clinician responses are reported as counts and percentages and as a mean normalized score on a 0--100 scale, where yes, partial, and no were assigned values of 100, 50, and 0, respectively. Confidence intervals were obtained from 10{,}000 case-level bootstrap resamples.

Cross-system comparisons were conducted for the three computational endpoints: model-assessed safety, instruction concreteness, and guideline alignment. The primary complete-case comparison included the 400 prescriptions with valid scores from all six systems. For each continuous endpoint, we applied a Friedman omnibus test followed, when significant, by paired Wilcoxon signed-rank tests comparing FITTER with each comparator. We report mean paired differences with 95\% confidence intervals from 10{,}000 paired case-level bootstrap resamples and matched-pairs rank-biserial correlations. For model-assessed safety, we applied Cochran's $Q$ followed, when significant, by exact paired McNemar tests, reporting paired risk differences and discordant-pair counts. Holm correction controlled multiplicity across the five FITTER--comparator tests within each endpoint. The cancer-cohort ablation used the same endpoint-specific paired procedures, with full FITTER as the reference and Holm correction across the three ablation comparisons. All tests were two-sided, and bootstrap resampling used a fixed seed.

\subsection{Implementation and reproducibility}

All generation-time language-model components in FITTER, including query planning and safety review, and all comparator and ablation generation calls used Qwen3.5-9B (\texttt{Qwen/\allowbreak Qwen3.5-9B}) as a common backbone. We selected this open-weight model because it could be deployed within the secure All of Us Researcher Workbench, keeping participant-level generation inputs inside the approved environment to protect privacy and comply with the program's restrictions on exporting participant-level data~\cite{allofus_privacy}. Its reported instruction-following and agent-task performance suited the multistep generation task, while the 9B variant fit the memory and throughput limits of the available computing environment~\cite{qwen2026qwen35}. We used greedy decoding (temperature~$0$; sampling disabled) for all systems and generated one final prescription per participant per system. Retrieval used the \texttt{bge-base-en-\allowbreak v1.5} encoder and FAISS index described in \cref{subsec:rag}.

\subsection{Ethical approval declarations}

This study analyzed de-identified data from the All of Us Research Program~\cite{all2019all}, accessed within the secure, cloud-based All of Us Researcher Workbench (Registered Tier, Curated Data Repository version~8). Participants provided informed consent for broad research use of their data at in-person enrollment sites or through the program's web-based portal, and enrollment and data collection are approved by the All of Us institutional review board~\cite{all2019all,allofus_irb}. Direct identifiers are removed and dates shifted; all researchers completed the required data-access training and signed the Data User Code of Conduct~\cite{allofus_privacy}. Because the analysis used only such records, it does not meet the All of Us IRB definition of human-subjects research and is not subject to the Common Rule (45 CFR part 46), and therefore required no additional IRB review~\cite{all2019all,commonrule}.

\section{Data availability}\label{sec12}
This study used data from the All of Us Research Program Registered Tier Dataset version 8, available to authorized users through the Researcher Workbench. Access to participant-level data is restricted and requires authorization under the All of Us Research Program data access policies, including institutional coverage under a Data Use and Registration Agreement, identity verification, completion of required research training, and acknowledgment of the Data User Code of Conduct. Participant-level data cannot be redistributed by the authors. All analyses of All of Us participant-level data were conducted within the secure Researcher Workbench environment in accordance with the program's data use policies.

\section{Code availability}\label{sec13}
The FITTER framework, including the five prescription agents and their prompts, the guideline retrieval and indexing pipeline, the medication knowledge-base construction, the safety-constraint mapping and deterministic audit and repair logic, the five comparator systems, and the ablation conditions, is publicly available at \url{https://github.com/carolli07/fittvp}. The repository contains no participant-level data. Code that executes inside the All of Us Researcher Workbench, including the cohort-extraction and profile-construction queries against the Registered Tier dataset, cannot be redistributed under All of Us Research Program policy. The guideline documents indexed by the retrieval component are third-party copyrighted publications and are not redistributed; the repository lists their citations and the passage-segmentation parameters required to rebuild the index.

\section{Acknowledgements}\label{sec14}
We gratefully acknowledge All of Us participants for their contributions, without whom this research would not have been possible. We also thank the National Institutes of Health's All of Us Research Program for making available the participant data examined in this study. This research did not receive any specific grant from funding agencies in the public, commercial, or not-for-profit sectors.

\section{Author contributions}\label{sec15}
X.L., Y.L., Y.W., and R.Y. conceived and designed the study. X.L., Y.W., and R.Y.  designed and implemented the FITTER framework, the comparator systems, and the ablation conditions, and performed the computational evaluation and statistical analysis. X.L. contributed to cohort construction from the All of Us Registered Tier dataset, data curation, and software implementation. F.W. and M.W. provided technical and infrastructure support for the study. Y.L., R.A.G., J.K., J.S., and S.A.B. provided clinical domain expertise for the five-agent workflow, the eligible activity set, and the safety-constraint mapping. Y.L., R.A.G., J.K., and J.S. conducted the structured clinician review of generated prescriptions. Y.L., J.S., and S.A.B. advised on cancer-survivorship and stroke-rehabilitation content. X.L. wrote the original draft. C.X. contributed to review and revision of the manuscript. R.Y., Y.L., and X.H. supervised the study. All authors reviewed, edited, and approved the final manuscript.

\section{Competing interests}\label{sec16}
S.A.B. is the lead author of the American Heart Association/American Stroke Association scientific statement on physical activity and exercise for stroke survivors~\cite{billinger2014physical}, one of the four documents comprising the guidance corpus from which FITTER retrieves and against which the guideline-alignment endpoint is computed. S.A.B. contributed clinical domain expertise to the design of the agent workflow but did not implement the framework, conduct the structured clinician review, or take part in the computational evaluation. The remaining authors declare no competing interests.

\newpage

\newpage
\bibliography{sn-bibliography}
\includepdf[pages=-]{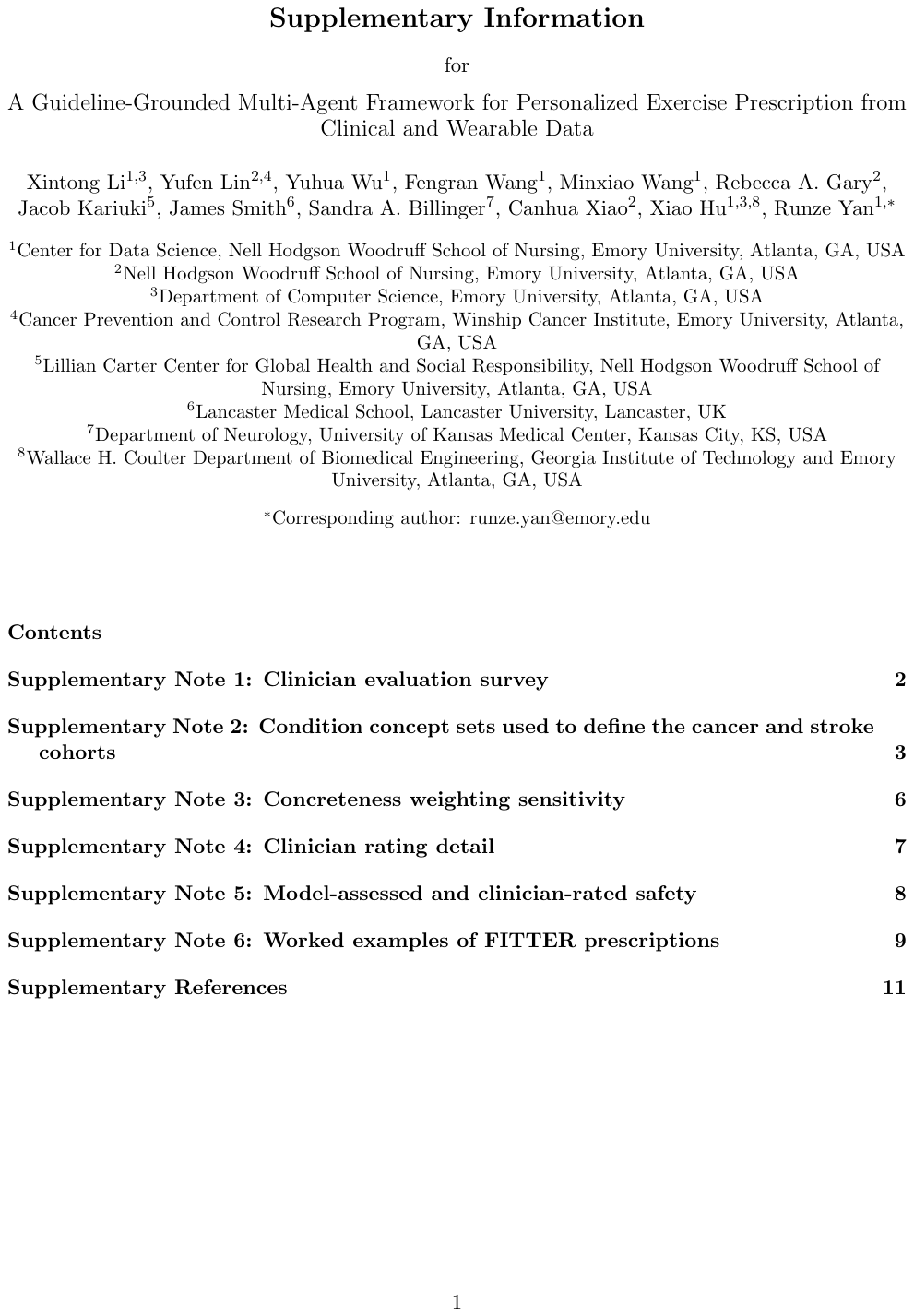}
\end{document}